\documentclass[letterpaper, 10 pt, conference]{ieeeconf}  

\IEEEoverridecommandlockouts                              

\usepackage{graphicx}
\usepackage{subfigure}
\usepackage{amsmath}
\usepackage{amssymb}
\usepackage{booktabs}
\usepackage{multirow}
\usepackage{multicol}
\usepackage{cite}
\usepackage[utf8]{inputenc}
\usepackage{pifont}
\usepackage[colorlinks,
            linkcolor=blue,
            anchorcolor=blue,
            citecolor=blue,
            urlcolor=blue]{hyperref}

\usepackage[left=0.75in, right=0.75in, top=0.76in, bottom=0.79in]{geometry}

\title{\LARGE \bf
MK-Pose: Category-Level Object Pose Estimation via Multimodal-Based Keypoint Learning
}

\author{Yifan Yang$^{1}$, Peili Song$^{1}$, Enfan Lan$^{1}$, Dong Liu$^{1}$ and Jingtai Liu$^{1*}$, Senior Member, IEEE
\thanks{This work is supported by the National Natural Science Foundation of China under Grant 62173189. }
\thanks{$^{1}$The authors are with the Institute of Robotics and Automatic Information System, Tianjin Key Laboratory of Intelligent Robotics, and also with TBI center, Nankai University, Tianjin 300350, China. 
        {\tt\small Emails: yangyifan@mail.nankai.edu.cn; peilisong@mail.nankai.edu.cn; lef@mail.nankai.edu.cn; dongliu@nankai.edu.cn; liujt@nankai.edu.cn.}}%
\thanks{*Corresponding author}
}

\begin{document}

\maketitle
\thispagestyle{empty}
\pagestyle{empty}

\begin{abstract}
Category-level object pose estimation, which predicts the pose of objects within a known category without prior knowledge of individual instances, is essential in applications like warehouse automation and manufacturing. Existing methods relying on RGB images or point cloud data often struggle with object occlusion and generalization across different instances and categories. This paper proposes a multimodal-based keypoint learning framework (MK-Pose) that integrates RGB images, point clouds, and category-level textual descriptions. The model uses a self-supervised keypoint detection module enhanced with attention-
based query generation, soft heatmap matching and graph-based relational modeling. Additionally, a graph-enhanced feature fusion module is designed to integrate local geometric information and global context. MK-Pose is evaluated on CAMERA25 and REAL275 dataset, and is further tested for cross-dataset capability on HouseCat6D dataset. The results demonstrate that MK-Pose outperforms existing state-of-the-art methods in both IoU and average precision without shape priors. Codes will be released at \href{https://github.com/yangyifanYYF/MK-Pose}{https://github.com/yangyifanYYF/MK-Pose}.

\end{abstract}

\section{Introduction}
In modern robotics, robotic arm grasping is a fundamental task, with applications spanning industries such as warehouse automation, manufacturing, and healthcare \cite{wei2024learning, yang2024attribute}. Robotic arms are tasked with perceiving objects in the environment, accurately localizing them, and performing grasping actions. As automation levels increase, robots need to interact with a wide variety of objects in dynamic and cluttered environments, where these objects can vary greatly in shape, size, and material. Therefore, obtaining precise pose information of these objects becomes crucial for successful and reliable grasping.

Object pose estimation involves determining the rotation and translation of an object relative to a reference frame. In our previous work \cite{10802046}, we have introduced a point cloud based symmetry-aware 6D pose estimation method, which deals with challenges such as rust, high reflectivity, and absent textures. However, instance-level pose estimation methods \cite{drost2010model, wang2019densefusion, peng2019pvnet, wang2020self6d, cai2022ove6d, duffhauss2023symfm6d, 10802046} rely on prior knowledge of specific object models, making them highly accurate but inflexible when dealing with novel objects. Real-world applications often require robots to interact with unseen objects and categories. This necessity has given rise to category-level object pose estimation, which aims to estimate the pose of objects within a given category without prior knowledge of their specific instances.

\begin{figure}[tb]	
	\centering
  \includegraphics[width=1\linewidth]{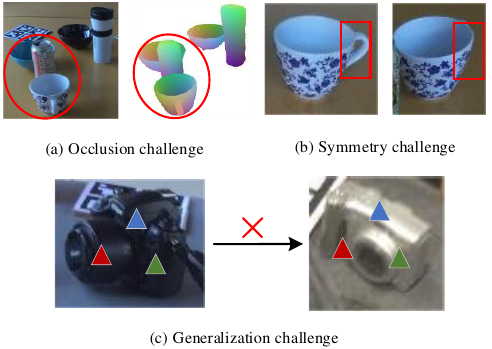}
	\caption{Three main challenges in category-level object pose estimation. (a) shows the issue of mutual occlusion among bowl, can, and mug; The mug in (b) is an asymmetric object when the handle is visible, and a symmetric object when the handle is not visible; In (c), when the type of camera changes, the original keypoints do not generalize well.}
    \label{challenge}
\end{figure}

Despite the significant advances in category-level pose estimation in previous work \cite{wang2019normalized, lin2021dualposenet, di2022gpv, liu2023prior, wang2023query6dof, lin2024instance, zhang2024lapose, tian2020shape, chen2021sgpa, lin2022sar, zhang2022rbp, lin2022category, zhang2024generative, zhang2024omni6dpose, you2022cppf, wang2024gs, 10802487}, there are still some challenges. Firstly, the majority of current approaches rely solely on RGB images or point cloud data without incorporating prior knowledge about object categories, which hinders the ability to generalize across diverse instances and recover the occluded object (Fig. \ref{challenge}(a)). Secondly, many keypoint-based methods require manually annotated keypoints for each object instance, which hampers generalization to unseen objects, as the learned keypoint extraction strategies may not be transferable, and acquiring precise keypoint annotations for every possible object instance is labor-intensive and impractical in large-scale applications (Fig. \ref{challenge}(c)). Thirdly, extracted features often fail to effectively integrate global shape information with fine-grained local details. Also, most existing approaches primarily handle infinite axial symmetry. The lack of a comprehensive, generalizable solution for handling diverse symmetry types leads to ambiguities and incorrect pose predictions (Fig. \ref{challenge}(b)).

In response to the above challenges, we propose a multimodal-based keypoint learning framework for category-level pose estimation (MK-Pose). MK-Pose processes RGB images, point clouds, and object category text information to generate a unified object representation. In the self-supervised keypoint detection module, the fused feature enhanced with positional encoding and global context is interacted with learnable queries via an attention mechanism to generate refined keypoint features. Keypoint positions are determined through a soft heatmap matching process, with a graph-based network refining the relationships between keypoints. In the graph-enhanced local\&global feature fusion module, relative positional encodings enhance spatial relationships, while a cross-attention mechanism integrates local details. Absolute positional encodings and a global feature vector are leveraged through self-attention to further refine keypoint features. Finally, for pose and size estimation, a specialized loss function accounts for object symmetry, considering predefined symmetric transformations and an infinite symmetry vector.

In summary, the main contributions of this work are as follows:

\begin{enumerate}
    \item We propose MK-Pose that processes multimodal input including RGB images, point clouds and texts, achieving excellent zero-shot performance when encountering previously unseen objects.
    \item We employ learnable queries and graph-based networks in MK-Pose, ensuring accurate and generalizable self-supervised keypoint detection.
    \item We evaluate MK-Pose on REAL275 and CAMERA25 dataset, and test it for cross-dataset capability on HouseCat6D dataset. The results shows that MK-Pose outperforms other state-of-the-art methods.
\end{enumerate}
\section{Related Work}
\subsection{Instance-level Object Pose Estimation}
Instance-level object pose estimation aims to recover the 6D pose of known object instances from data acquired by visual sensors, given the CAD model of the object. Early methods primarily rely on feature-based matching approaches, such as 3D model alignment using features like SIFT and ORB. \cite{drost2010model} Additionally, ICP (Iterative Closest Point) and its variants are widely used to optimize poses in 3D point clouds. \cite{hinterstoisser2013model} However, these methods tend to fail in cases with low texture or significant lighting changes. With the development of deep learning, researchers begin using neural networks to directly regress 6D poses. DenseFusion \cite{wang2019densefusion} introduces a strategy of fusing RGB and point cloud features, enhancing robustness against occlusions and complex backgrounds. PVNet \cite{peng2019pvnet} performed keypoint regression and employed RANSAC post-processing to improve pose estimation stability. In recent years, researchers have also started exploring ways to reduce dependence on large-scale annotated datasets. For instance, Self6D \cite{wang2020self6d} used self-supervised learning to improve generalization. OVE6D \cite{cai2022ove6d} estimates the 6D object pose from a single depth image and the target object mask. SyMFM6D \cite{duffhauss2023symfm6d} a symmetry-aware multi-directional fusion approach for multi-view 6D object pose estimation, enhancing accuracy and robustness in complex scenarios. Despite the impressive performance of instance-level 6D pose estimation in known object scenarios, it requires a pre-existing 3D CAD model of the object, making it unsuitable for scenes with novel objects.

\subsection{Category-level Object Pose Estimation}
Category-level object pose estimation aims to estimate the pose of objects from the same category but different instances, without the need for a complete 3D CAD model in advance. This task is more challenging due to significant variations in object shape, size, and texture. The first batch of category-level 6D pose estimation methods borrowed ideas from instance-level keypoint detection. For example, NOCS \cite{wang2019normalized} introduced a Normalized Object Coordinate Space to recover 6D poses by predicting the normalized coordinates of the object. Inspired by NOCS, multiple works \cite{lin2021dualposenet, di2022gpv, liu2023prior, wang2023query6dof, lin2024instance, zhang2024lapose} have been proposed to better address the problem of category-level pose estimation. To better adapt to shape variations across different instances, some researchers also introduce shape modeling-based approaches \cite{tian2020shape, chen2021sgpa, lin2022sar, zhang2022rbp, lin2022category}. A series of methods represented by GenPose \cite{zhang2024generative, zhang2024omni6dpose} leverage score-based diffusion models, framing pose estimation as a multi-hypothesis problem, which allow for more robust predictions by exploring different possible solutions. Some researchers also train their models using only simulation data and test them on real-world datasets, demonstrating the algorithm's cross-domain capability \cite{you2022cppf, wang2024gs, 10802487}. 
\section{Method}
\begin{figure*}[tb]
  \centering
  \includegraphics[width=1\linewidth]{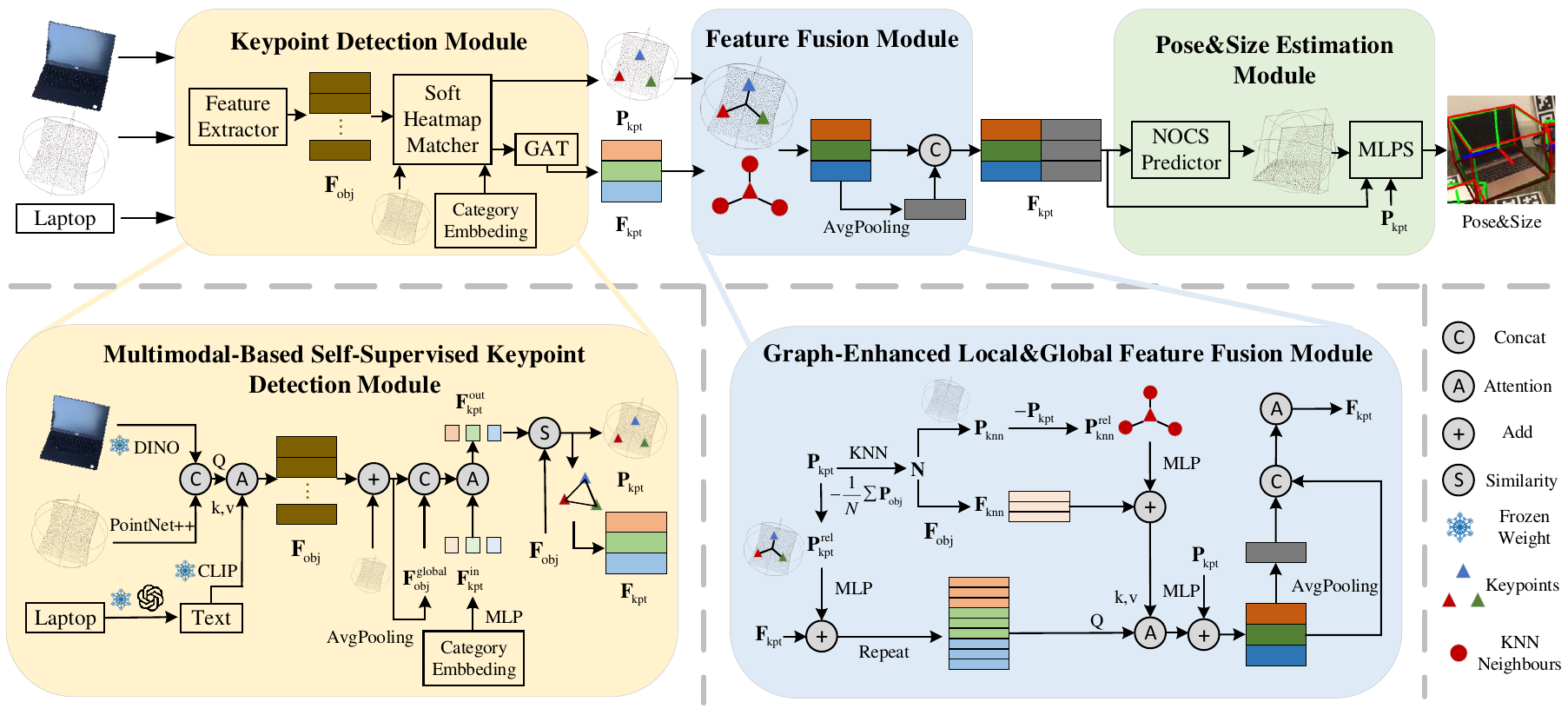}
  \caption{Architecture of MK-Pose. 
The framework takes multimodal inputs, including RGB images, point clouds, and text. It estimates the object's pose and size through multimodal-based self-supervised keypoint detection module and the graph-enhanced local\&global feature fusion module.}
  \label{network}
\end{figure*}
\subsection{Architecture Overview and Problem Formulation}
As shown in Fig. \ref{network}, the input of MK-Pose consists of RGB image \(\mathbf{I}\), point cloud \(\mathbf{P}\), and object category information \(c\). The output is rotation matrix $\mathbf{R}$, translation vector $\mathbf{t}$, size $\mathbf{s}$. The pipeline of the proposed method consists of three major modules. 1) The keypoint detection module encodes and fuses multimodal inputs, and obtains  keypoint positions and features via self-supervised approach. 2) The feature fusion module refines keypoint features by fusing local and global features. 3) The pose\&size estimation module finally estimates the pose and size of the target object.

In keypoint detection module, \(\mathbf{I}\), \(\mathbf{P}\), and \(c\) are encoded into \(\mathbf{F}_{\text{RGB}}\), \(\mathbf{F}_{\text{PC}}\), and \(\mathbf{F}_{\text{text}}\), which are fused to form \(\mathbf{F}_{\text{obj}}\). $\mathbf{F}_{\text{obj}}$ incorporates positional encoding and global information to form $\mathbf{F}_{\text{obj}}'$. Learnable keypoint queries \(\mathbf{F}_{\text{kpt}}^{\text{in}}\) interact with \(\mathbf{F}_{\text{obj}}'\) via attention to produce \(\mathbf{F}_{\text{kpt}}^{\text{out}}\). Keypoint positions \(\mathbf{P}_{\text{kpt}}\) and features \(\mathbf{F}_{\text{kpt}}\) are derived through soft heatmap matching and a Graph Attention Network (GAN).

In feature fusion module, relative position encodings \(\mathbf{P}_{\text{kpt}}^{\text{rel}}\) and \(\mathbf{P}_{\text{knn}}^{\text{rel}}\) are added to \(\mathbf{F}_\text{kpt}\) and k-nearest neighbor features \(\mathbf{F}_\text{knn}\). Using a cross-attention mechanism, we fuse local information to update \(\mathbf{F}_\text{kpt}\), which is further refined by adding absolute position encoding \(\mathbf{P}_{\text{kpt}}\) and concatenating the global feature \(\mathbf{F}_{\text{kpt}}^{\text{global}}\).

Additionally, we design a general loss function \( \mathcal{L}_{\text{ps}} \) that considers symmetry. By leveraging a set of symmetric rotation matrices \( \mathcal{R}_{\text{S}} \) and an infinite symmetry vector \( \mathbf{v} \), it can be applied to both finitely and infinitely symmetric objects.

\subsection{Multimodal-Based Self-Supervised Keypoint Detection Module}
RGB features convey appearance details such as color and texture, while depth features contain geometric structure. However, they may misalign—e.g., objects with similar textures may have different shapes, and vice versa. Text descriptions provide a shared semantic space that can help bridge this gap. Additionally, occlusions complicate pose estimation with RGB-D alone, while text remains invariant, offering a stable reference. Aligning RGB-D features with text enhances the model’s robustness and consistency in category-level representation.

\begin{figure}[tb]
  \centering
  \includegraphics[width=1\linewidth]{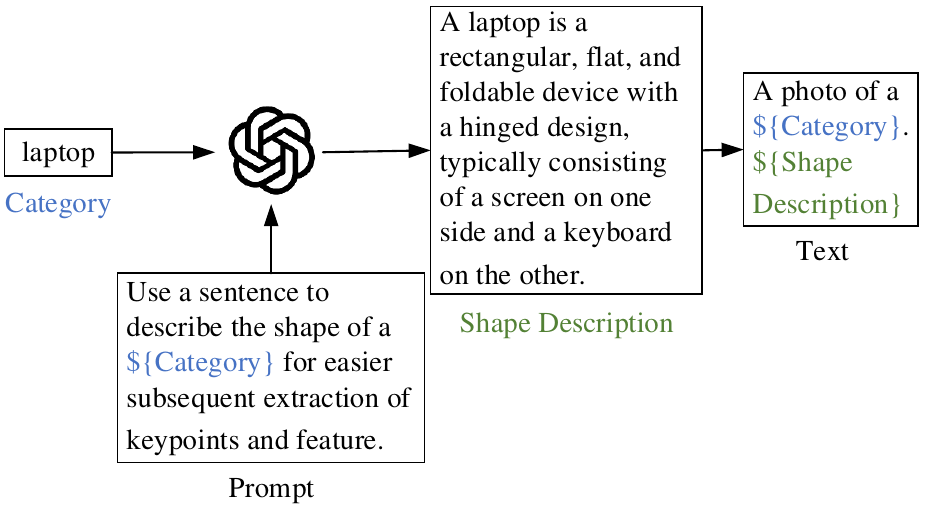}
  \caption{Text generation process based on object categories. 
Given an object category (e.g., "laptop"), a large language model is prompted to generate a textual description of its shape. This shape description, which details the object's geometric and structural characteristics, is then incorporated into a formatted text prompt for downstream feature extraction. }
  \label{gpt}
\end{figure}

Given an input RGB image \(\mathbf{I} \in \mathbb{R}^{H \times W \times 3}\), we utilize the DINOv2-ViT-S/14 model \(\mathcal{F}_{\text{RGB}}\) \cite{oquab2023dinov2} to extract visual features
\begin{equation}
     \mathbf{F}_{\text{RGB}} = \mathcal{F}_{\text{RGB}}(\mathbf{I}) \in \mathbb{R}^{N \times d_1}.
\end{equation}
The depth information \(\mathbf{D} \in \mathbb{R}^{H \times W}\) is first transformed into a point cloud \(\mathbf{P} \in \mathbb{R}^{N \times 3}\) using camera intrinsics. We apply PointNet++ \(\mathcal{F}_{\text{PC}}\) \cite{qi2017pointnet++} to extract geometric features
\begin{equation}
     \mathbf{F}_{\text{PC}} = \mathcal{F}_{\text{PC}}(\mathbf{P}) \in \mathbb{R}^{N \times d_2},
\end{equation}
where $N$ is the number of sampled points. We leverage category-level textual information to provide semantic context. As is shown in Fig. \ref{gpt}, given a category name \(c\), we use a large language model (LLM) GPT-4o to generate a detailed textual description, and use the CLIP text encoder \(\mathcal{F}_{\text{text}}\) \cite{radford2021learning} to obtain the text feature representation
\begin{equation}
     \mathbf{F}_{\text{text}} = \mathcal{F}_{\text{text}}(\text{LLM}(c)) \in \mathbb{R}^{N \times d}.
\end{equation}
The fused \(\mathbf{F}_{\text{obj}}\in \mathbb{R}^{N \times d}\) is finally obtained by a cross-attention mechanism
\begin{equation}
\begin{aligned}
    &\mathbf{Q} = W_Q \text{cat}(\mathbf{F}_{\text{RGB}},\mathbf{F}_{\text{PC}}), \\
&\mathbf{K} = W_K \mathbf{F}_{\text{text}}, \quad
\mathbf{V} = W_V \mathbf{F}_{\text{text}}, \\
    &\mathbf{F}_{\text{obj}}=\text{softmax} \left( \frac{\mathbf{Q} \mathbf{K}^\top}{\sqrt{d}} \right) \mathbf{V}.
\end{aligned}
\end{equation}

Keypoints are detected using a combination of attention-based query generation, soft heatmap matching and graph-based relational modeling.

To enhance position and global awareness, the raw input features \(\mathbf{F}_{\text{obj}}\) are first combined with positional encodings derived from point cloud coordinates and global features from average pooling
\begin{equation}
     \mathbf{F}_{\text{obj}}' = \text{MLP}([\mathbf{F}_{\text{obj}}+\text{MLP}(\mathbf{P}), \frac{1}{N} \sum_{i=1}^{N} (\mathbf{F}_{\text{obj}}+\text{MLP}(\mathbf{P}))_{i,:}]).
\end{equation}
Each learnable keypoint feature query \(\mathbf{F}_{\text{kpt}}^{\text{in}} \in \mathbb{R}^{N_{\text{kpt}} \times d}\) generated from  category embeddings attends to the object feature \(\mathbf{F}_{\text{obj}}'\) using an attention mechanism, so that we can get $\mathbf{F}_{\text{kpt}}^{\text{out}} \in \mathbb{R}^{N_{\text{kpt}} \times d}$ guided by category information
\begin{equation}
     \mathbf{F}_{\text{kpt}}^{\text{out}} = \text{Attn}(\mathbf{F}_{\text{kpt}}^{\text{in}}, \mathbf{F}_{\text{obj}}').
\end{equation}
The keypoint coordinates $\mathbf{P}_{\text{kpt}} \in \mathbb{R}^{N_{\text{kpt}} \times 3}$ and keypoint features $\mathbf{F}_{\text{kpt}} \in \mathbb{R}^{N_{\text{kpt}} \times d}$ are obtained by weighted sum
\begin{equation}
     \mathbf{P}_{\text{kpt}} = \mathbf{H}\mathbf{P}, \quad\mathbf{F}_{\text{kpt}} = \mathbf{H}\mathbf{F}_{\text{obj}},
\end{equation}
where $\mathbf{H} \in \mathbb{R}^{N_{\text{kpt}} \times N}$ is a heatmap predicted by the cosine similarity between $\mathbf{F}_{\text{kpt}}^{\text{out}}$ and $\mathbf{F}_{\text{obj}}$. It is defined as
\begin{equation}
     \mathbf{H} = \text{Softmax}(\frac{S}{\tau}),
     \label{H}
\end{equation}
and the cosine similarity $S$ is calculated as
\begin{equation}
     S_{ij} = \frac{\mathbf{F}_{\text{kpt}, i} \cdot \mathbf{F}_{\text{obj}, j}}{\|\mathbf{F}_{\text{kpt}, i}\|_2 \|\mathbf{F}_{\text{obj}, j}\|_2 + \epsilon}, \, i \in [1, N_{\text{kpt}}], \, j \in [1, N],
     \label{s}
\end{equation}
where $\tau$ is a temperature parameter, and $\epsilon$ is a small positive constant.

To model structural relationships between keypoints, a Graph Attention Network (GAT) is applied. We define a graph \( G = (\mathcal{V}, \mathcal{E}) \), where \( \mathcal{V} = \mathbf{F}_{\text{kpt}} \) represents the keypoints as nodes. \( \mathcal{E} \subseteq \mathcal{V} \times \mathcal{V} \) represents the edges between keypoints, which is defined based on fully connected topology. The network update process is
\begin{equation}
     h_i' = \sigma \left( \sum_{j \in \mathbf{N}_i \cup \{i\}} \alpha_{ij} \mathbf{W} h_j \right),
\end{equation}
where \( h_i  \) represents the feature vector of node \( i \). \( \mathbf{W} \) is a learnable weight matrix. \( \mathbf{N}_i \) denotes the neighbors of node \( i \). \( \sigma \) is a non-linear activation function. \( \alpha_{ij} \) is the attention coefficient, determining the importance of node \( j \) to node \( i \). Instead of using fixed adjacency weights, we use a learnable attention mechanism to dynamically determine the importance of each neighbor, and finally obtain the updated features $\mathbf{F}_{\text{kpt}}$. 
\subsection{Graph-Enhanced Local\&Global Feature Fusion Module}
Accurately capturing both local geometric relationships and global contextual information is crucial for enhancing the quality of keypoint feature representations. However, achieving this balance in feature processing remains a challenge, particularly when dealing with spatially distributed data such as keypoints and point clouds. 

In the first stage of feature fusion module, to extract local information, a K-Nearest Neighbors (KNN) graph is built based on the sorted pairwise Euclidean distance for each keypoint. Let $\mathbf{N}_i = \{ j_1, j_2, \dots, j_k \} $ represent the indices of the \( k \)-nearest neighbors for the $i$-th keypoint. 
\begin{equation}
     \mathbf{N}_i = {\arg\min}_{\{j_1, j_2, \dots, j_k\}} \sum_{m=1}^{k} D_{i, j_m}
\end{equation}
These nearest neighbor points \( \mathbf{P}_{j_m} \) (where \( m \in \{1, 2, \dots, k\} \)) correspond to the smallest \( k \) distances \( D_{i,j_m} \), where
\begin{equation}
    D_{i,j} = \|\mathbf{P}_{\text{kpt}, i} - \mathbf{P}_{\text{obj}, j}\|_2.
    \label{Dij}
\end{equation}
The coordinates $\mathbf{P}_{\text{knn}, i} \in \mathbb{R}^{k \times 3}$ and features $\mathbf{F}_{\text{knn}, i} \in \mathbb{R}^{k \times d}$ of the $k$ nearest neighbors corresponding to the $i$-th keypoint can be indexed as
\begin{equation}
\begin{aligned}
    \mathbf{P}_{\text{knn}, i} &= \left\{ \mathbf{P}_{j_1}, \mathbf{P}_{j_2}, \dots, \mathbf{P}_{j_k} \right\},\\
    \mathbf{F}_{\text{knn}, i} &= \left\{ \mathbf{F}_{\text{obj}, j_1}, \mathbf{F}_{\text{obj}, j_2}, \dots, \mathbf{F}_{\text{obj}, j_k} \right\}.
\end{aligned}
\end{equation}
To better capture the geometric information such as the shape of the object, we incorporate relative position encoding into $\mathbf{F}_{\text{kpt}} \in \mathbb{R}^{N_{kpt} \times d}$ and $\mathbf{F}_{\text{knn}} \in \mathbb{R}^{N_{kpt} \times k \times d}$
\begin{equation}
\begin{aligned}
    \mathbf{F}_{\text{kpt}} &= \mathbf{F}_{\text{kpt}} + \text{MLP}(\mathbf{P}_{\text{kpt}}^{\text{rel}}),\\
    \mathbf{F}_{\text{knn}} &= \mathbf{F}_{\text{knn}} + \text{MLP}(\mathbf{P}_{\text{knn}}^{\text{rel}}).
\end{aligned}
\end{equation}
where $\mathbf{P}_{\text{kpt}}^{\text{rel}}$ is keypoints' position relative to the object center, while $\mathbf{P}_{\text{knn}}^{\text{rel}}$ is the neighbors' position relative to the corresponding keypoint
\begin{equation}
\begin{aligned}
    \mathbf{P}_{\text{kpt}}^{\text{rel}} &= \mathbf{P}_{\text{kpt}} - \frac{1}{N} \sum_{j=1}^{N} \mathbf{P}_{j},\\
    \mathbf{P}_{\text{knn}}^{\text{rel}} &= \mathbf{P}_{\text{knn}} - \mathbf{P}_{\text{kpt}}.
\end{aligned}
\end{equation}
By repetition and rearrangement, the dimension of $\mathbf{F}_{\text{kpt}}$ and $\mathbf{F}_{\text{knn}}$ is transformed into $\mathbb{R}^{( N_{kpt} \cdot k ) \times d}$. The keypoint features $\mathbf{F}_{\text{kpt}}$ serve as queries, while neighbor features $\mathbf{F}_{\text{knn}}$ act as keys and values in a multi-head attention mechanism
\begin{equation}
     \mathbf{F}_{\text{kpt}} = \text{Attn}(\mathbf{F}_{\text{kpt}}, \mathbf{F}_{\text{knn}}).
\end{equation}

In the second stage, to fuse global information, the global feature of the keypoints, denoted as $\mathbf{F}_{\text{kpt}}^{\text{global}}$, is obtained by calculating the mean of the keypoint features with the addition of the absolute positional encoding of the keypoints
\begin{equation}
    \mathbf{F}_{\text{kpt}} = \mathbf{F}_{\text{kpt}} + \text{MLP}(\mathbf{P}_{\text{kpt}}),
\end{equation}
\begin{equation}
    \mathbf{F}_{\text{kpt}}^{\text{global}} = \frac{1}{N_{\text{kpt}}} \sum_{i=1}^{N_{\text{kpt}}} \mathbf{F}_{\text{kpt}, i}.
\end{equation}
This global feature is broadcasted to all keypoints and fused with the local keypoint features via an MLP, and then the feature representation is enhanced using a self-attention mechanism.
\subsection{Symmetry-Aware Pose\&Size Estimation Module and Loss}
In pose\&size estimation module, we predict the keypoint NOCS coordinates from the obtained keypoint features via MLPs and attention, and finally regress the object's rotation matrix $\mathbf{R}$, translation vector $\mathbf{t}$, size $\mathbf{s}$ following \cite{lin2022category}. To address the issue of symmetry ambiguity in the pose estimation task, based on our previous work \cite{10802046}, we propose a general symmetry-aware loss function $\mathcal{L}_{\text{ps}}$ for pose and size that can uniformly handle objects with no symmetry, finite symmetry, and infinite symmetry
\begin{equation}
    \begin{aligned}
        & \mathcal{L}_{\text{ps}} =  \min_{\mathbf{R}_{\text{S}} \in \mathcal{R}_{\text{S}}}  \left\| \mathbf{R}^{\text{gt}}\mathbf{v}\mathbf{R}_{\text{S}} - \mathbf{R}^{\text{pred}}\mathbf{v} \right\|_2 \\
        & + \left\| \mathbf{t}^{\text{gt}} - \mathbf{t}^{\text{pred}} \right\|_2 
        + \left\| \mathbf{s}^{\text{gt}} - \mathbf{s}^{\text{pred}} \right\|_2.
    \end{aligned}
    \label{pose_loss}
\end{equation}
For each finitely symmetric object (e.g., a prism), a set of symmetric rotation matrices \( \mathcal{R}_{\text{S}} \) can be computed, while for other objects, \( \mathcal{R}_{\text{S}} \) contains only the identity matrix. Objects with infinite rotational symmetry around an axis (e.g., a cylinder) are abstracted as directed line segments located on the symmetry axis. $\mathbf{v} \in \{0, 1\}^{3\times1}$ is the vector representing the infinite symmetry of the object
\begin{equation}
    \mathbf{v} =
\begin{bmatrix}
1 - I(\text{x}) \\
1 - I(\text{y}) \\
1 - I(\text{z})
\end{bmatrix},
    \label{v}
\end{equation}
where
\begin{equation}
I(a) =
\begin{cases}
1, & \text{if infinitely symmetric about the } a \text{ axis} \\
0, & \text{otherwise}
\end{cases}
.
\end{equation}
\subsection{Overall Loss Function}
The overall loss function can be represented as follows
\begin{equation}
    \mathcal{L}_{\text{all}} = \lambda_1 \mathcal{L}_{\text{ps}} + \lambda_2 \mathcal{L}_{\text{cd}} + \lambda_3 \mathcal{L}_{\text{div}} + \lambda_4 \mathcal{L}_{\text{rec}} + \lambda_5 \mathcal{L}_{\Delta} + \lambda_6 \mathcal{L}_{\text{nocs}},
\end{equation}
where $\lambda_1, \lambda_2, \dots, \lambda_6$ are weights for each corresponding loss function, and $\mathcal{L}_{\text{nocs}}$ is from \cite{lin2022category}.

The loss functions \(\mathcal{L}_{\text{cd}}\) and \(\mathcal{L}_{\text{div}}\) constrain keypoint detection in a self-supervised manner, without requiring true keypoint locations. \(\mathcal{L}_{\text{cd}}\) ensures each predicted keypoint is near at least one observed point, while \(\mathcal{L}_{\text{div}}\) promotes diversity, preventing keypoints from collapsing.
\begin{equation}
    \mathcal{L}_{\text{cd}} = \frac{1}{N_{\text{kpt}}} \sum_{j=1}^{N_{\text{kpt}}}\min_{i\in [1, N]}\|\mathbf{P}_i-\mathbf{P}_{\text{kpt}, j}\|_2,
\end{equation}
\begin{equation}
    \mathcal{L}_{\text{div}} = \frac{1}{N_{\text{kpt}}(N_{\text{kpt}}-1)} \sum_{i=1}^{N_{\text{kpt}}}\sum_{j=1, j \neq i}^{N_{\text{kpt}}}L_{ij},
\end{equation}
where a threshold $T$ and a linear mapping are applied to constrain the loss
\begin{equation}
    L_{ij} = 1-\frac{\min(\|\mathbf{P}_{\text{kpt},i}-\mathbf{P}_{\text{kpt},j}\|_2, T)}{T}.
\end{equation}

Our self-supervised method for 3D keypoint detection ensures geometric consistency by using MLPs to predict local offsets for keypoints, generating a dense 3D reconstruction. Each keypoint expands into a local point cloud with \( N_{\text{p}} \) points, forming the reconstructed set \(\mathbf{P}_{\text{rec}} \in \mathbb{R}^{3 \times (N_{\text{kpt}} N_{\text{p}})}\) and offsets \(\Delta\mathbf{P}\). The reconstruction loss \(\mathcal{L}_{\text{rec}}\) enforces consistency between \(\mathbf{P}_{\text{rec}}\) and the observed point cloud \(\mathbf{P}\)
\begin{equation}
\begin{aligned}
    \mathcal{L}_{\text{rec}} = &\frac{1}{N} \sum_{j=1}^{N}\min_{i \in [1, N_{\text{kpt}}N_{\text{p}}]}\|\mathbf{P}_i-\mathbf{P}_{\text{rec}, j}\|_2+\\
    &\frac{1}{N_{\text{kpt}}N_{\text{p}}} \sum_{j=1}^{N_{\text{kpt}}N_{\text{p}}}\min_{i \in [1, N]}\|\mathbf{P}_{\text{rec}, i}-\mathbf{P}_j\|_2.
\end{aligned}
\end{equation}
$\mathcal{L}_{\Delta}$ ensures that there will be no significant deviation during reconstruction
\begin{equation}
     \mathcal{L}_{\Delta} = \mathbb{E} \left[ \left\| \Delta\mathbf{P} \right\|_2 \right].
\end{equation}

\section{Experiment}
\subsection{Experimental Setup}
\noindent\textbf{Implementation details \quad}In the data preparation phase, we use a mixture of synthetic and real data as the training set. For RGB images, we crop and scale them to 224 × 224. For depth data, we convert them into point clouds based on camera intrinsic parameters, with the number of sampled points \( N = 1024 \). In the training phase, the feature dimensions are \( d_1 = d_2 = 128 \) and \( d = 256 \), the number of keypoints \( N_{\text{kpt}} = 96 \), the temperature parameter $\tau = 0.1$, $\epsilon = 10^{-7}$, the number of neighbors \( k = 16 \), and the diversity threshold \( T = 0.01 \). The weights for the loss functions are set as \( \lambda_1 = 0.3 \), \( \lambda_2 = 2.0 \), \( \lambda_3 = 10.0 \), \( \lambda_4 = 15.0 \), \( \lambda_5 = 1.0 \), and \( \lambda_6 = 2.0 \). The batch size is set to 32 during training. In the testing phase, similar to previous work, the method first uses MaskRCNN \cite{he2017mask} to segment the observed scene and obtain instance masks. All training and testing are conducted on a single RTX3090Ti GPU.
\begin{table*}[tb]
	\renewcommand\arraystretch{1.2}
    \setlength{\abovecaptionskip}{0pt}    
    \setlength{\belowcaptionskip}{5pt}
	\centering
	\caption{Comparison of different methods on REAL275 dataset}
    \setlength{\tabcolsep}{14pt}
	\resizebox{1\textwidth}{!}{
		\begin{tabular}{lccccccc}
\toprule[1.5pt]
Method & Use of Shape Priors & IoU$_{50}$ & IoU$_{75}$ & 5°/2cm & 10°/2cm & 10°/5cm \\ \hline
NOCS \cite{wang2019normalized}          & \ding{55} & 78.0  & 30.1 & 7.2  & 13.8  & 25.2  \\ 
DualPoseNet \cite{lin2021dualposenet}   & \ding{55} & 79.8  & 62.2 & 29.3 & 50.0  & 66.8  \\ 
GPV-Pose \cite{di2022gpv}       & \ding{55} & ---   & 64.4 & 32.0 & ---   & 73.3  \\ 
IST-Net \cite{liu2023prior}       & \ding{55} & 82.5  & 76.6 & 47.5 & 72.1  & 80.5  \\ 
Query6DoF \cite{wang2023query6dof}     & \ding{55} & 82.5  & 76.1 & 49.0 & 68.7  & 83.0  \\ 
LaPose \cite{zhang2024lapose}     & \ding{55} & 47.9  & 15.8 & --- & 37.4  & 57.4  \\ 
GenPose \cite{zhang2024generative}   & \ding{55} & ---  & --- & 52.1 & 72.4  & 84.0  \\ 
AG-Pose \cite{lin2024instance}         & \ding{55} & 83.7  & 79.5 & 54.7 & 74.7  & 83.1  \\
\hline
SPD \cite{tian2020shape}           & \ding{51} & 77.3  & 53.2 & 19.3 & 43.2  & 54.1  \\ 
SGPA \cite{chen2021sgpa}           & \ding{51} & 80.1  & 61.9 & 35.9 & 61.3  & 70.7  \\ 
SAR-Net \cite{lin2022sar}       & \ding{51} & 79.3  & 62.4 & 31.6 & 50.3  & 68.3  \\ 
RBP-Pose \cite{zhang2022rbp}      & \ding{51} & ---   & 67.8 & 38.2 & 63.1  & 79.1  \\ 
DPDN \cite{lin2022category}          & \ding{51} & 83.4  & 70.6 & 46.0 & 70.4  & 78.4  \\ 
 \hline
MK-Pose(Ours)          & \ding{55} & \textbf{84.0}  & \textbf{80.3} & \textbf{60.8} &\textbf{78.0}  & \textbf{84.6}  \\ 
\bottomrule[1.5pt]
\end{tabular}}
	\label{real}
\end{table*}
\begin{table}[tb]
	\renewcommand\arraystretch{1.2}
 \setlength{\abovecaptionskip}{0pt}    
        \setlength{\belowcaptionskip}{5pt}
	\centering
	\caption{Comparison of different methods on CAMERA25 dataset}
	\resizebox{0.49\textwidth}{!}{
		\begin{tabular}{lcccccc}
\toprule[1.5pt]
Method & IoU$_{50}$ & IoU$_{75}$ & 5°/2cm & 10°/2cm & 10°/5cm \\ \hline
NOCS \cite{wang2019normalized}          & 83.9 & 69.5 & 32.3 & 48.2 & 64.4 \\ 
DualPoseNet \cite{lin2021dualposenet}   & 92.4 & 86.4 & 64.7 & 77.2 & 84.7 \\ 
GPV-Pose \cite{di2022gpv}       & 93.4 & 88.3 & 72.1 & --- & 89 \\ 
Query6DoF \cite{wang2023query6dof}     & 91.9 & 88.1 & 78.0 & 83.9 & 90 \\ 
LaPose \cite{zhang2024lapose}     & 49.4  & 14.1 & --- & 42.4  & 73.1  \\ 
GenPose \cite{zhang2024generative}   & ---  & --- & \textbf{79.9} & 84.6  & 89.4  \\ 
AG-Pose \cite{lin2024instance}         & 93.8  & 91.3 & 77.8 & 85.5  & 91.6  \\
\hline
SPD \cite{tian2020shape}           & 93.2 & 83.1 & 54.3 & 73.3 & 81.5 \\ 
SGPA \cite{chen2021sgpa}           & 93.2 & 88.1 & 70.7 & 82.7 & 88.4 \\ 
SAR-Net \cite{lin2022sar}       & 86.8 & 79 & 66.7 & 75.3 & 80.3  \\ 
RBP-Pose \cite{zhang2022rbp}      & 93.1 & 89 & 73.5 & 82.1 & 89.5  \\ \hline
MK-Pose(Ours) & \textbf{94.1} & \textbf{92.2} & 77.9 & \textbf{86.1} & \textbf{91.7}\\
 \bottomrule[1.5pt]
\end{tabular}}
	\label{camera}
\end{table}

\noindent\textbf{Datasets \quad}Following previous works on category-level pose estimation \cite{wang2019normalized, lin2021dualposenet, di2022gpv, liu2023prior, wang2023query6dof, lin2024instance, zhang2024lapose, tian2020shape, chen2021sgpa, lin2022sar, zhang2022rbp, lin2022category, zhang2024generative, you2022cppf, wang2024gs, 10802487}, We evaluate our approach on two widely used datasets: CAMERA25 and REAL275 \cite{wang2019normalized}. CAMERA25 is a synthetic dataset derived from the ShapeNet dataset, consisting of six object categories (bottle, can, bowl, laptop, camera, and mug) with a total of 1,085 instances rendered under various poses and lighting conditions. Among them, 184 instances are designated for validation. In total, 300K composited images are generated, with 25K reserved for validation. REAL275 is a real-world dataset collected using a structure sensor in cluttered indoor environments. It consists of 7,000 images from 13 distinct scenes. Out of these, 2,750 images from six scenes are set aside for validation, including three novel instances per category that were not present in the training set. By utilizing both datasets, we assess our method’s ability to generalize from synthetic to real-world data. To demonstrate the cross-dataset capability of our method, we also choose to test it on the HouseCat6D dataset \cite{jung2024housecat6d}. This test set consists of five test scenes (3k frames) containing objects from 10 household categories, including photometrically challenging objects such as glass and cutlery, with occlusions.

\noindent\textbf{Evaluation metrics \quad}Following NOCS\cite{wang2019normalized}, 
for symmetric object categories like bottles, bowls, and cans, we permit the predicted 3D bounding box to rotate freely around the object's vertical axis. For the mug, we consider it a symmetric object when the handle is not visible. To assess 3D detection, we employ the intersection over union (IoU) metric with a 25\%, 50\% and 75\% threshold, which can be represented as IoU$_{25}$, IoU$_{50}$ and IoU$_{75}$ respectively. For pose estimation, we present the average precision of object instances where the translation error is less than m cm and the rotation error is below $n$°, which is represented as $n$°/$m$ cm in the results tables. 

\subsection{Comparison Results}
\begin{figure}[tb]
  \centering
  \includegraphics[width=1\linewidth]{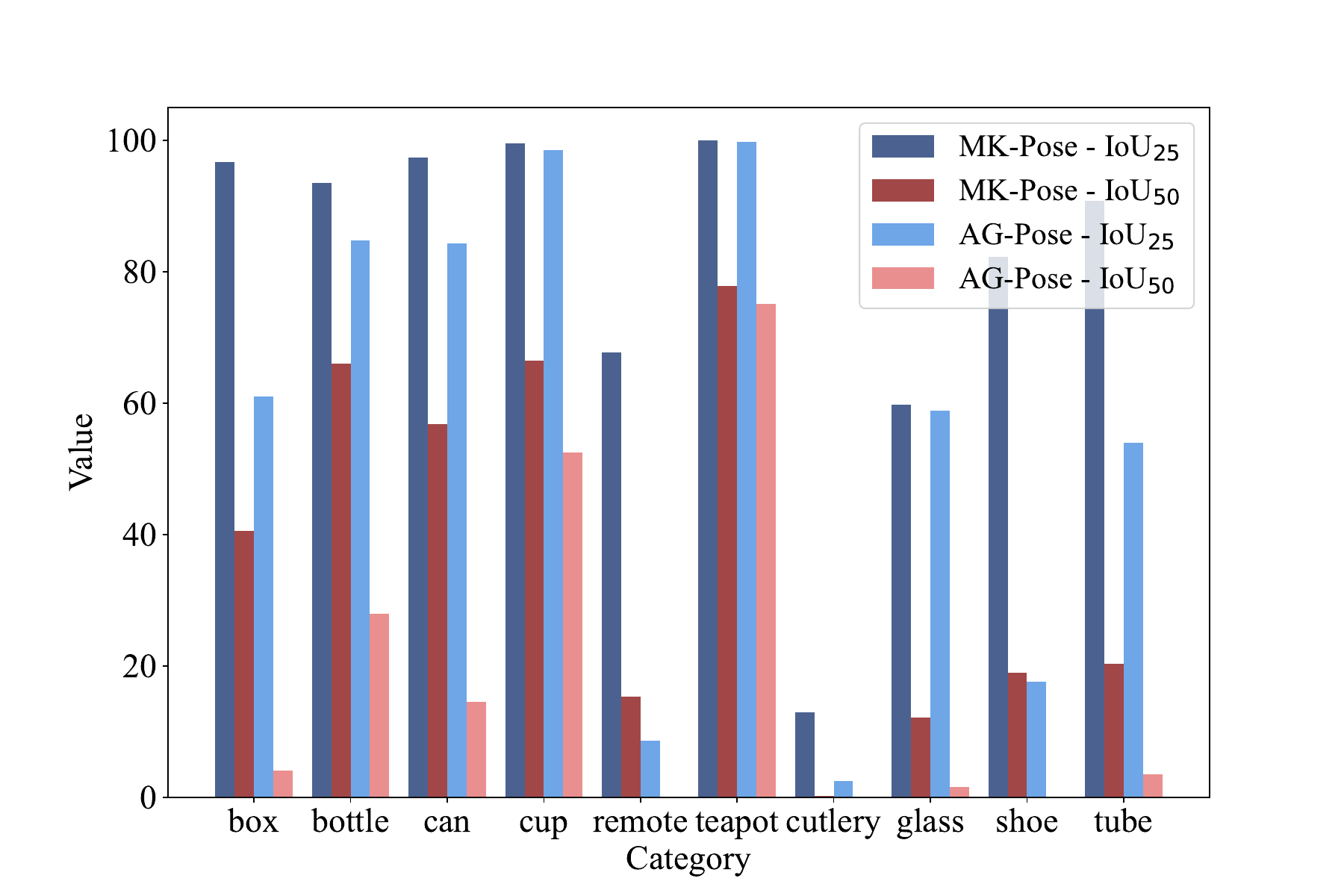}
  \caption{Comparison on HouseCat6D dataset. The bar chart shows the IoU$_{25}$ and IoU$_{50}$ of MK-Pose and AG-Pose for different categories of objects.}
  \label{HouseCat}
\end{figure}
\begin{figure*}[tb]
  \centering
  \includegraphics[width=1\linewidth]{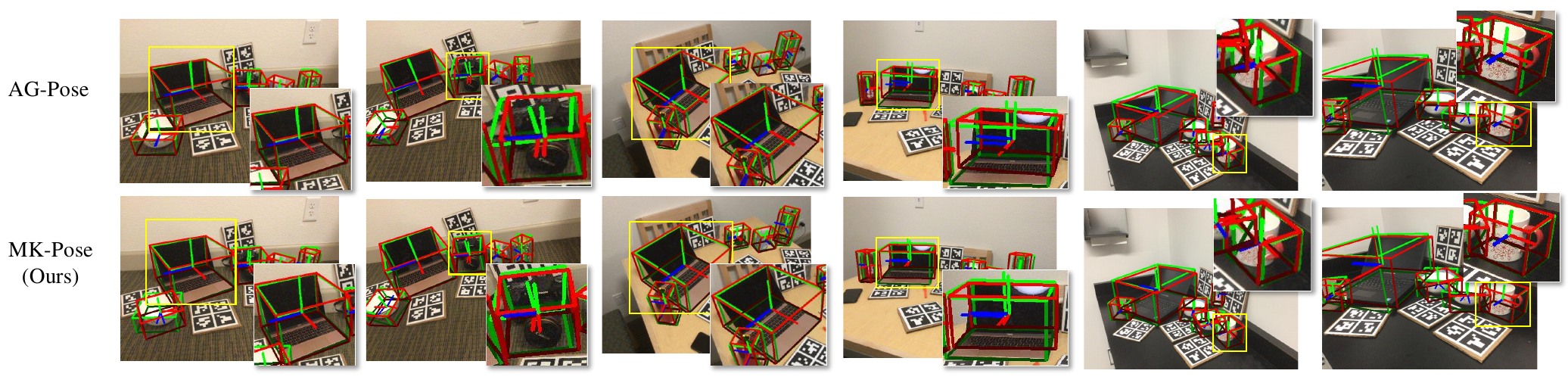}
  \caption{Visualization results comparison of our method and the  of the most advanced methods in current category-level pose estimation. The red bounding box represents the predicted result, while the green bounding box represents the ground truth. Bowls, cans, and bottles exhibit infinite rotational symmetry around the y-axis, so any bounding box errors around the y-axis are considered correct.}
  \label{viz}
\end{figure*}
MK-Pose is compared with existing state-of-the-art methods to demonstrate its effectiveness. Tab. \ref{real} presents a comparison of different methods on the REAL275 dataset, evaluating their performance based on IoU and pose estimation average precision at various error thresholds. The proposed method, MK-Pose, achieves state-of-the-art performance without using shape priors. Among the shape-prior-free methods, GenPose \cite{zhang2024generative} and AG-Pose \cite{lin2024instance}  have demonstrated strong performance. However, MK-Pose surpasses both in multiple aspects, highlighting its effectiveness in pose estimation. For example, at 5°/2cm, MK-Pose achieves 60.8\%, exceeding GenPose’s 52.1\% by 8.7\% and AG-Pose’s 54.7\% by 6.1\%. Methods utilizing shape priors generally perform well, but MK-Pose achieves superior results despite not relying on them. Even compared to the best shape-prior-based method, DPDN \cite{lin2022category}, MK-Pose performs better, with an improvement of 14.8\% at 5°/2cm. Tab. \ref{camera} presents a comparison of different methods for category-level pose estimation on the synthetic CAMERA25 dataset. Despite not using shape priors, MK-Pose surpasses all other methods in IoU50, IoU75, 10°/2cm and 10°/5cm, and achieves comparable results to Gen-Pose in 5°/2cm, demonstrating that it can achieve high-precision results without the need for additional shape priors.

To evaluate the cross-dataset generalization ability of our approach, we trained the model on REAL275 and CAMERA25, and tested it on the HouseCat6D dataset. The results are presented in Fig. \ref{HouseCat}, comparing our method, MK-Pose, with AG-Pose across IoU$_{25}$ and IoU$_{50}$. Our method significantly outperforms AG-Pose in all categories. For categories not present in the training set, such as box, shoe, and tube, MK-Pose demonstrates a considerable advantage. Specifically, MK-Pose achieves an IoU$_{50}$ of 37.5\%, nearly 1.9× improvement over AG-Pose (19.7\%). Similarly, for IoU$_{25}$, MK-Pose attains 80.1\%, surpassing AG-Pose (63.6\%) by 16.5\%. This indicates that when faced with objects never seen in the training set, MK-Pose demonstrates a strong advantage in a zero-shot setting. These results demonstrate the robustness and superior generalization ability of our method, allowing it to accurately handle novel objects without requiring prior exposure during training.


\subsection{Ablation Study}
Tab. \ref{ablation study} presents an ablation study on category-level object pose estimation, comparing full MK-Pose with five ablated versions, where "w/o" means without, i.e., removing the specific module only. The variant "w/o text" omits text feature generated from CLIP, leading to a decrease in performance, which highlights the importance of text-based features in restoring occluded objects and providing a shared semantic space. "w/o sym" uses plain L$_2$ loss instead of symmetry-aware pose\&size loss function, indicating the necessity of symmetry perception for pose estimation. The absence of feature fusion module in "w/o feature fusion" leads to a further reduction in performance, highlighting the value of a balanced integration of local geometric information and global context to keypoint features. The "w/o reconstructor" setting results indicates that the reconstruction module refines pose estimation by enforcing geometric consistency. Removing global features in keypoint detection module ("w/o global") suggests that global contextual information plays a key role in stable keypoint predictions.
\begin{table}[tb]
	\renewcommand\arraystretch{1.2}
    \setlength{\abovecaptionskip}{0pt}    
    \setlength{\belowcaptionskip}{5pt}
	\centering
	\caption{Results of ablation study}
	\resizebox{0.45\textwidth}{!}{
		\begin{tabular}{cccc}
			\toprule[1.5pt]
			Method & 5°/2cm & 10°/2cm & 10°/5cm \\ \hline
        w/o text & 53.3 & 72.4 & 83.2 \\ 
        w/o sym & 58.4 & 72.9 & 82.0 \\ 
        w/o feature fusion & 54.1 & 73.4 & 83.7 \\
        w/o reconstructor & 55.2 & 72.8 & 81.8\\
        w/o global & 57.8 & 74.8 & 82.6 \\\hline
        MK-Pose(Ours) & \textbf{60.8} &\textbf{78.0}  & \textbf{84.6} \\
			\bottomrule[1.5pt]
	   \end{tabular}}
	\label{ablation study}
\end{table}

\subsection{Visualization}
To evaluate the performance of our category-level object pose estimation approach, we conduct qualitative visualization experiments, comparing our method with the most advanced method AG-Pose \cite{lin2024instance}. As shown in Fig. \ref{viz}, our method demonstrates superior precision and robustness when dealing with objects that are occluded or with complex and unclear textures. While AG-Pose struggles with objects that are partially hidden, often predicting incorrect orientations or misalignments, our method successfully captures plausible poses. This suggests that our approach effectively leverages contextual and geometric cues beyond visible keypoints, leading to more reliable pose predictions in cluttered environments. Furthermore, our method handles symmetric objects in a more practical and efficient manner. AG-Pose tends to expend considerable effort in regressing a pose that is closer to the ground truth in terms of a specific canonical orientation. However, for symmetric objects, multiple equivalent poses exist, and forcing the model to predict a particular one can introduce unnecessary errors. In contrast, our approach naturally converges to one of the valid poses, leading to a more stable and consistent performance across different instances. 
\section{Conclusion}
In this paper, we propose MK-Pose, a multimodal keypoint learning framework for category-level object pose estimation. By integrating RGB images, point clouds, and category-level textual descriptions, MK-Pose constructs a unified object representation to enhance accuracy and generalization ability. A self-supervised keypoint detection module obtains features using learnable queries and a graph-based network, while a graph-enhanced feature fusion module improves robustness against occlusions and intra-category variations. Additionally, a symmetry-aware general pose\&size loss function is introduced to handle both finite and infinite symmetry, boosting performance on symmetric objects.  
Extensive experiments on REAL275 and CAMERA25 show that
MK-Pose achieves state-of-the-art performance, outperforming shape-prior-free methods and
its shape-prior-based counterparts.
Cross-dataset tests on HouseCat6D demonstrate its strong zero-shot generalization ability to unseen objects, showcasing its effectiveness in real-world applications.
\bibliography{references}
\bibliographystyle{IEEEtran}

\end{document}